\documentclass[conference]{IEEEtran}
\IEEEoverridecommandlockouts
\usepackage{cite}
\usepackage{amsmath,amssymb,amsfonts}
\usepackage{algorithmic}
\usepackage{graphicx}
\usepackage{textcomp}
\usepackage{mathptmx}
\usepackage[dvipsnames]{xcolor}
\usepackage{mathtools}
\usepackage{booktabs}
\usepackage{mdframed}
\PassOptionsToPackage{bookmarks=false}{hyperref}
\usepackage[most]{tcolorbox}

\graphicspath{ {./images/} }
\def\BibTeX{{\rm B\kern-.05em{\sc i\kern-.025em b}\kern-.08em
    T\kern-.1667em\lower.7ex\hbox{E}\kern-.125emX}}

\newtcolorbox{shadedcvbox}[1][]{enhanced jigsaw,
  colback=white!85!gray,
  coltext={black},
  boxrule=0pt,
  arc=3mm,
  auto outer arc,
  boxsep=3pt,
  left=4pt,
  right=2pt,
  bottom=6pt,
  top=6pt,
  #1}
  
\begin{document}

\title{Clustering-based Automatic Construction of Legal Entity Knowledge Base from Contracts}

\author{\IEEEauthorblockN{1\textsuperscript{st} Fuqi Song}
\IEEEauthorblockA{\textit{Data Science, Hyperlex} \\
\\
Paris, France\\
fsong@hyperlex.ai}
\and
\IEEEauthorblockN{2\textsuperscript{nd} Éric de la Clergerie}
\IEEEauthorblockA{\textit{Data Science, Hyperlex} \\
\textit{Alpage, INRIA}\\
Paris, France \\
Eric.De\_La\_Clergerie@inria.fr}}

\maketitle

\makeatletter
\def\ps@IEEEtitlepagestyle{%
  \def\@oddfoot{\mycopyrightnotice}%
  \def\@evenfoot{}%
}
\def\mycopyrightnotice{%
  {\footnotesize 978-1-7281-6251-5/20/\$31.00 ©2020 IEEE\hfill}
  \gdef\mycopyrightnotice{}
}

\begin{abstract}
In contract analysis and contract automation, a Knowledge Base (\textit{KB}) of legal entities is fundamental for performing tasks such as contract verification, contract generation and contract analytic. However, such a knowledge base does not always exist nor can be produced in a short time. In this paper, we propose a clustering-based approach to automatically generate a reliable knowledge base of legal entities from given contracts without any supplemental references. The proposed method is robust to different types of errors produced by pre-processing such as Optical Character Recognition (\textit{OCR}) and Named Entity Recognition (\textit{NER}), as well as editing errors such as typos. We evaluate our method on a dataset that consists of 800 real contracts with various qualities from 15 clients. Compared to the collected ground-truth data, our method is able to recall 84\% of the knowledge.
\end{abstract}


\vspace{2mm}
\section{Introduction}
\label{sec-introductino}
In the life cycle of contract management, the clauses of party declaration play an important role. They declare the legal entities involved in an engagement and the legal responsibilities of each party. Many contract analysis and automation tasks are based on these clauses and legal entities such as:

\begin{itemize}{}{}
\item \textit{Contract verification}: check inconsistencies and errors that are present in contracts;
\item \textit{Contract generation}: generate and complete automatically certain clauses and fields specified by contract templates;
\item \textit{Contract analytics}: analyze statistically the partnerships and provide advanced business insights for customers.
\end{itemize}

A reliable knowledge base of legal entities is critical to perform the above tasks. However such a knowledge base is not always publicly available, nor can the clients provide such data easily in a short time. In this paper, we propose an unsupervised clustering-based approach to automatically extract a such reliable knowledge base from contracts. Three preliminary document processing steps are required: OCR (\textit{Optical Character Recognition}), NER (\textit{Named Entity Recognition})~\cite{Nadeau2007,Yadav2019} and entity aggregation~\cite{Pons2006,Fortunato2010}. 

The rest of our paper is organized as follows: Section~\ref{sec-problem} explains the basic concepts relevant to legal entities and discusses the problems meet in legal entity extraction; Section~\ref{sec-related-works} studies the related works in the domains of entity linking and ontology population; Section~\ref{sec-ap} presents the proposed clustering-based approach; Section~\ref{sec-eval} demonstrates the evaluation of our method on a dataset that consists of 800 real contracts and Section~\ref{sec-conc} draws some conclusions of this paper and extends the future works.

\vspace{2mm}
\section{Problem Statement}
\label{sec-problem}
According to the Cambridge Dictionary\footnote{https://dictionary.cambridge.org/dictionary/english/legal-entity}, a legal entity is defined as \textit{a company or organization that has legal rights and responsibilities}. Concretely, in a contract, a legal entity is characterized by a set of information. The following example is a typical way to describe a legal entity in a French contract. The corporate name is usually used to represent the legal entity and a few attributes (in bold) are associated with it. In this paper, we use the term \textit{basic entity} to refer to these attributes and the term \textit{legal entity} to refer to the structure formed by these basic entities. This paper focuses on the basic entities with the following roles: \textit{corporate name, nature, capital, registration number, registration city, headquarter address and legal representative}.

\begin{shadedcvbox}
\begin{quote}
 \textit{\ldots, \textbf{Hyperlex}, \textbf{société à actions simplifiées} au capital \textbf{2040,78 euros}, dont le siège social est \textbf{sis 12 rue Anselme 93400 Saint-Ouen}, immatriculée au registre du commerce et des sociétés de \textbf{Bobigny} sous le numéro \textbf{832 146 237}, \ldots}   
\end{quote}
\end{shadedcvbox}

\begin{figure*}[h]
\centering
\includegraphics[width=16cm]{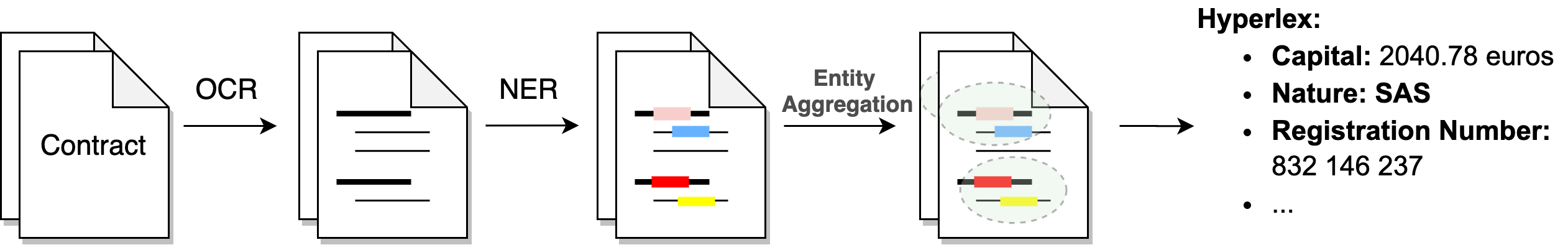}
\caption{Pipeline for extracting a raw legal entity from a single contract}
\label{fig:01}
\end{figure*}

For legal entity extraction, generally we apply three key processing steps as illustrated in Figure~\ref{fig:01}: 1) OCR for recognizing text from documents, 2) NER for extracting (basic) named entities from text and 3) entity aggregation for regrouping the basic entities that belong to the same legal entity. Assuming that these steps work perfectly without errors, we could get a clean legal entity from a contract. However, in practice, none of these steps works perfectly, for instance, OCR works poorly on scanned and old documents; NER performs variously on different contexts and different types of entities; not to mention editing errors such as wrong information sources and typos. Given a single contract, usually only partial basic entities could be extracted and often containing errors. In the following examples, two basic entities are processed with errors: \textit{corporate name} and \textit{legal representative}:

\begin{itemize}
 \item ``\textit{\textbf{John Doe} is\_representative\_of \textbf{Compamy AbcD}}": appeared three times (typos in corporate name);
 \item ``\textit{\textbf{Jean Doe} is\_representative\_of \textbf{Company ABC}}": appeared once (wrong name of the legal representative);
 \item ``\textit{\textbf{NA} is\_representative\_of \textbf{Company ABC}}": appeared once (missing legal representative);
 \item ``\textit{\textbf{John Doe} is\_representative\_of \textbf{Company ABC Ltd}}": appeared twice (variation of corporate name),
\end{itemize}
where the ground-truth is: ``\textit{\textbf{John Doe} is\_representative\_of \textbf{Company ABC}}". In order to extract the correct legal entity from such noisy data, we propose two key operations: 
\begin{enumerate}
  \item we refer a group of aggregated basic entities as \textit{raw legal entity} and use $g_{ij}$ to denote it. A raw legal entity may contain partial information with possible errors. The first operation is to aggregate these raw legal entities $ \{g_{11},\ g_{12},\ \dots , \ g_{21},\ g_{22},\ \dots, g_{ij},\ \dots\}$ into groups $g_1=\{g_{11},\ g_{12},\ \dots\}$, $g_2=\{g_{21},\ g_{22},\ \dots\},\ \dots $, each group presenting potentially the same legal entity;
  
  \item in each group $g_{i}$, find a representative value for each basic entity. In this way, we select the representatives for all types of basic entities and form the final legal entity for group $g_i$. 
\end{enumerate}

\begin{figure*}[h]
\centering
\includegraphics[width=17.2cm]{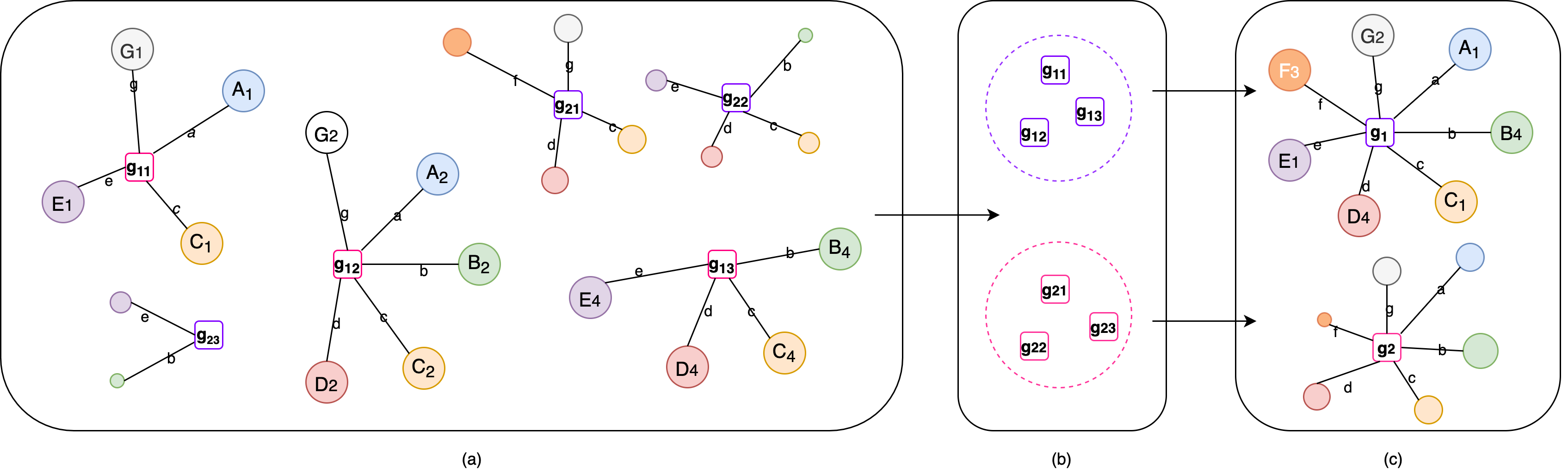}
\caption{Schema of general problem: the nodes with same color represent the basic entities with the same role (denoted also by the edge name), (a) raw legal entities with partial information and variations of basic entities, (b) grouped raw legal entities that present potentially the same legal entity and (c) the expected legal entity with complete information and representative value for each basic entity.}
\label{fig:02}
\end{figure*}

Figure~\ref{fig:02} illustrates visually the two operations. Different sub-graphs on the left (a) represent the raw legal entities $g_{ij}$ extracted from a contract base. The graphs in middle (b) present the grouped raw legal entities and on the right (c) is the expected legal entities with complete correct attributes.

\vspace{2mm}
\section{Related Works}
\label{sec-related-works}
We categorize the related works into two domains: Named Entity Linking (NEL) and Ontology Population (OP). NEL assigns a unique identity to named entities as an entry in a structured knowledge base~\cite{Hachey2013}. In our case, we group basic entities that belong to the same role and link them to the representative entity, i.e. unique identity. We don't link them to
an external KB such as Wikipedia used in~\cite{Hachey2013}. Many works have been published about NEL, Shen~\cite{Shen2015} gave a comprehensive review of different techniques and applications of NEL. Raiman~\cite{Raiman2018} stands as the current state-of-the-art (SOTA) in Cross-lingual Entity Linking for WikiDisamb30 and TAC KBP 2010 datasets. They constructed a deep type system and applied as a constraint of a neural network to keep the symbolic structure in the outputs. 

An ontology serves as an \textit{explicit specification of a conceptualization}~\cite{Gruber1993} and is generally used to express knowledge. Ontology population is \textit{the process of inserting concept and relation instances into an existing ontology}~\cite{Petasis2011}. We can regard a legal entity as an ontology and its basic entities as the properties. The purpose is to populate different raw legal entities into an empty ontology with a common structure. Ontology population is well studied in the domain of biology and medicine due to their needs of conceptualization for large sets of complex concepts. Petasis\cite{Petasis2011} discussed different research issues addressed in ontology population, and they also compared BOEMIE\cite{Fragkou2008} to other ontology extraction tools. Fareh\cite{Fareh2013} presented a similarity-based approach for ontology population based on measurements of three aspects: terminological, structural and semantic. 

Regarding relevant works in legal industry, Natural Language Processing (\textit{NLP}) and machine learning techniques are frequently adopted for: 1) extracting information from unstructured data, such as named entities\cite{Bruckschen2010} from legal texts and legal document types\cite{Wei2019}, and then 2) establishing links with existing KB. 
The European project MIREL\cite{Robaldo2018} studied and applied many NLP and ML techniques for extracting information from legal texts, such as privacy agreements, and connecting these extracted information to two knowledge bases: DAPRECO\cite{Robaldo2019} and PrOnto\cite{Palmirani2018}.

Most of the above-mentioned works explored establishing links between pieces of information (e.g. named entities from text) and entries in an existing knowledge base. However, often, such a knowledge base doesn't always exist nor easily accessible. In contract analysis, the clients do not often maintain their own knowledge base of legal entities, and the access to database of legal entities is not always free (e.g. in France). This is the motivation of our work on this paper. In the next section, we explain how our approach constructs such a KB in a autonomous way.

\vspace{2mm}
\section{Affinity Propagation Clustering}
\label{sec-ap}
Affinity Propagation (\textbf{AP})~\cite{Frey2007} clustering is an approach based on sending (real-valued) messages between pairs of examples until a high-quality set of exemplars and corresponding clusters gradually emerge. The messages represent the suitability for one example to be the representative of the other one. Updating occurs iteratively until convergence, at which point the final exemplars are chosen, and then the final clustering is obtained. In this way, exemplars are chosen by examples if they are similar enough to many other examples and chosen by many examples to be their representative.

\def\esim{\textit{$\mathrm{basic\_entity\_sim}$}}
\def\gsim{\textit{$\mathrm{legal\_entity\_sim}$}}

To apply AP clustering, we need to compute a similarity matrix that indicates the degree of similarity between each pair of elements. In our approach, we compute respectively the similarity matrix for basic entities and legal entities. 

For basic entities, we use a hybrid function $\esim$ that combines char-level (Sequence Matcher~\cite{SequenceMatcher}) and token-level (Jaccard Index~\cite{jaccard1901distribution}) string metrics to capture maximum information between two basic entities $e_1$ and $e_2$. 

\vspace{-0.5cm}
\begin{multline}
\esim (e_1, e_2) =  \max (\mathrm{sequence\_matcher}(e_1, e_2),\\ \mathrm{jaccard\_index}(e_1, e_2))
\end{multline}

As illustrated in Figure~\ref{fig:02}, a potential legal entity is denoted as a group of basic entities $e$ (e.g. $A_1$) with role $r$ (e.g. \textit{a}). For legal entity-level similarity \gsim, we: 1) calculate the similarity of pairwise basic entities ($e_1$,  $e_2$) that share the same role $r$ in groups $g_1$ and $g_2$; 2) assign weight $w_r$ to each role, which is assigned empirically according to the importance of the roles, e.g. the corporate name has a greater $w_r$ since it is a critical for distinguishing two legal entities, and 3) compute similarity between $g_1$ and $g_2$ using weighed sum.

\vspace{-0.5cm}
\begin{multline}
\gsim(g_1,g_2) = \Sigma_{(r,e_1)\in g_1, (r,e_2) \in g_2, w_r \in W} w_r \\ \cdot \esim(e_1,e_2) 
\end{multline}

\begin{figure*}[h]
\centering
\includegraphics[width=16.2cm]{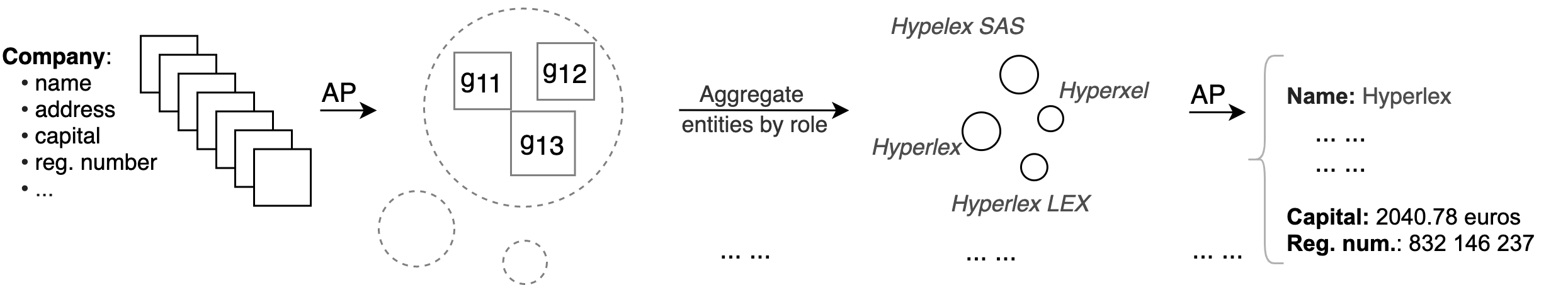}
\caption{Pipeline using Affinity Propagation to generate final legal entity from raw legal entities}
\label{fig:03}
\end{figure*}

Demonstrated in Figure~\ref{fig:03}, first we cluster the raw legal entities extracted from contract into clusters of groups using similarity matrix obtained by $\gsim(g_1,g_2)$. Then we aggregate the basic entities with the same roles in each cluster to find their exemplars using similarity matrix calculated by $\esim(e_1, e_2) $. In Figure~\ref{fig:03}, we show an example of the role of \textit{corporate name}. In the complete pipeline, we apply the same operations for all the required roles of entities mentioned in Section \ref{sec-problem}. Additionally, in practice, legal entities generated with less than three raw legal entities are removed since we regard them as unreliable.

\section{Evaluations and Discussions}
\label{sec-eval}
We use a dataset\footnote{The dataset is not accessible to public due to reasons of confidentiality} consisting of 800 contracts from 15 clients with ground-truth data collected on Infogreffe\footnote{Information of the Registries of the Commercial Courts, https://www.infogreffe.fr/} for French companies and on companies' websites for those of the other countries. Each contract declares several legal entities and is pre-processed by the pipeline described in Figure~\ref{fig:01}. 

In total, 572 reference legal entities are generated, most of them (90\%) have a cluster size smaller than three. Cluster size refers to the number of raw legal entities used in forming the final legal entity. This is due to the phenomenon that a client signed contracts with many different customers, thus this client himself appears much more frequently than the others. After eliminating those small clusters, we obtained 51 legal entities that are referred by the party declaration clauses in more than 50\% of the contracts.

We use two metrics to evaluate the performance of our method: 1) accuracy of key information extraction $\%key$ and 2) accuracy of general information extraction $\%all$. In a legal entity, the corporate name and the registration number are the two most important pieces of information. The first metric mentioned above evaluates the capability for capturing key information. The second metric evaluates the general extraction ability for all required information. Formally, for a generated legal entity:

\begin{itemize}
\item \textbf{$ \%key $} = 1 if both corporate name and registration number are correctly generated, otherwise 0 
\item \textbf{$\%all $} = \textit{the number of correctly generated basic entities} divided by \textit{the number of all expected basic entities}.
\end{itemize}

Table \ref{table:1} lists the average accuracy of the 51 generated legal entities. Overall, we obtained 80\% accuracy for key information and 84\% for all information in the legal entities that we produced fully automatically. These scores decrease as the size of clusters decreases: when the size is greater than 15, we generate 100\% key information and 91\% of all information, while 68\% and 78\% respectively when the size is smaller than 7. More examples benefits obviously to get better results. We investigated the legal entities with cluster size smaller than 3, the quality depends greatly on the pre-processing steps, meaning if we improve the accuracy of NER and entity aggregation, the performance of our method can be further improved. 

By analyzing other errors, we observe that 40\% of mismatched entities are due to the mismatch of headquarter addresses. This is due to the fact that many clients often use the address of their offices in their contacts instead of the legally registered one. Another 35\% mismatched entities are the capitals of companies that change regularly. The rest 25\% errors relate to the preliminary document processing steps and insufficient samples. 

\begin{table}
\begin{center}
\caption{Accuracy \textit{\%key} and \textit{\%all} by different ranges of cluster sizes}
\vspace{-3mm}
\label{table:1}
\begin{tabular}{ c|c|c|c|c } 
 \toprule
  & \textbf{all} & \textbf{size $>$ 15} & \textbf{7 $<$ size $\leq$ 15} & \textbf{3 $<$ size $\leq$ 7} \\
 \midrule
\textbf{\%key} & 80\% & 100\% & 87\% & 68\% \\
\textbf{\%all} & 84\% & 91\% & 87\% & 78\% \\
\textbf{Nb. of samples} & 51 & 11 & 15 & 25 \\
 \bottomrule
\end{tabular}
\end{center}
\vspace{-0.9cm}
\end{table}

\section{Concluding Remarks}
\label{sec-conc}

We presented a clustering-based automatic approach for legal entity base construction without relying on external knowledge base. The extracted knowledge base contributes considerably to different tasks in contract analysis and contract automation. Currently, our approach performs better on relatively static information with more data samples, and relies on a good pipeline for contract preliminary processing. 

Although our approach was tested on French contracts, it is language independent and can be applied to other languages with an adequate language specific pre-processing (OCR and NER). It is also applicable more broadly to other types of knowledge construction using the same principles. Based on our evaluation results, we plan on focusing on two main aspects in future work: 

\begin{itemize}{}{}
    \item quantify the reliability of generated legal entities using supplementary information provided by OCR and NER in the pre-processing phase in order to increase the generation capacity as the majority of the generated legal entities are currently discarded due to weak reliability;
    \item manage evolving information such as addresses, capitals and legal representative using approaches based on statistics and versioning in order to make the knowledge base more reliable.
\end{itemize}

\section*{Acknowledgment}
\vspace{-0.1cm}
We thank all the members of Data Science Team at Hyperlex for discussions and comments that improved this manuscript.

\bibliography{reference}
\vspace{-0.5cm}
\bibliographystyle{ieeetr}

\end{document}